# Companion Agents: A Table-Information Mining Paradigm for Text-to-SQL

**Jiahui Chen**[*†], **Lei Fu**[†], **Jian Cui**[‡], **Yu Lei, Zhenning Dong**

Amap, Alibaba Group

{ qianyan.cjh, arley.fl, cj87619, yu.lei, zhenning.dong}@alibaba-inc.com

## Abstract

Large-scale Text-to-SQL benchmarks such as BIRD typically assume complete and accurate database annotations as well as readily available external knowledge, which fails to reflect common industrial settings where annotations are missing, incomplete, or erroneous. This mismatch substantially limits the real-world applicability of state-of-the-art (SOTA) Text-to-SQL systems. To bridge this gap, we explore a database-centric approach that leverages intrinsic, fine-grained information residing in relational databases to construct missing evidence and improve Text-to-SQL accuracy under annotation-scarce conditions. Our key hypothesis is that when a query requires multi-step reasoning over extensive table information, existing methods often struggle to reliably identify and utilize the truly relevant knowledge. We therefore propose to "cache" query-relevant knowledge on the database side in advance, so that it can be selectively activated at inference time. Based on this idea, we introduce Companion Agents (CA), a new Text-to-SQL paradigm that incorporates a group of agents accompanying database schemas to proactively mine and consolidate hidden inter-table relations, value-domain distributions, statistical regularities, and latent semantic cues before query generation. Experiments on BIRD under the fully missing evidence setting show that CA recovers +4.49 / +4.37 / +14.13 execution accuracy points on RSL-SQL / CHESS / DAIL-SQL, respectively, with larger gains on the Challenging subset (+9.65 / +7.58 / +16.71). These improvements stem from CA's automatic database-side mining and evidence construction, suggesting a practical path toward industrial-grade Text-to-SQL deployment without reliance on human-curated evidence.

**Keywords:** Text-to-SQL; Large Language Models; Companion Agents; Query Routing; Database-side Knowledge Caching

## 1. Introduction

As digital transformation accelerates across industries, vast amounts of business data are accumulated in relational databases [1]. However, the ability to query and analyze such data is still gated by SQL expertise, creating a major bottleneck for non-technical users. Text-to-SQL, which translates natural language questions into executable SQL queries, has become a critical

---

[*] Work done during the internship at Amap, Alibaba Group.
[†] Equal contribution.
[‡] Corresponding author

component for democratizing data access and enabling conversational analytics [2].

Despite this promise, a substantial gap remains between academic benchmarks and real-world databases. Public benchmarks often assume that (i) every table is accompanied by complete and accurate schema descriptions; and (ii) critical signals such as foreign keys, enumerations, and domain rules are explicitly provided as supplementary knowledge [3]. For example, although BIRD queries are generally more complex than those in Spider, BIRD includes essential external evidence (e.g., formulas, enumerations, domain rules) in an evidence field [4]. As shown in Figure 1, as the evidence field becomes partially or fully missing, execution accuracy drops nearly linearly, with the most severe degradation on challenging queries where evidence frequently encodes indispensable additional information.

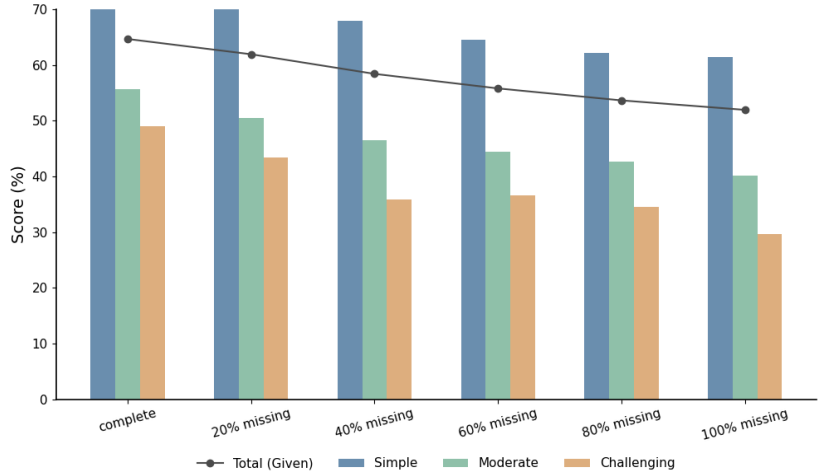

Figure 1. Execution accuracy on BIRD under different levels of evidence missingness.

Moreover, many competitive Text-to-SQL systems are evaluated under this "evidence-available" assumption, including robust schema-linking approaches (e.g., RSL-SQL [5]), chain-of-thought prompting methods (e.g., ACT-SQL [6]), retrieval-and-verification pipelines (e.g., CHESS [7]), alignment-based hallucination mitigation (e.g., TA-SQL [8]), and fine-tuning oriented frameworks (e.g., CodeS [9], PET-SQL [10]). As observed by SEED [11], most top-ranked approaches on the BIRD leaderboard rely heavily on the evidence field, exacerbating the gap between benchmark performance and deployable performance. Therefore, developing Text-to-SQL methods that can automatically supplement missing knowledge without human-labeled evidence is crucial.

We introduce Companion Agents (CA) to address Text-to-SQL degradation caused by missing evidence. CA not only enriches database-side knowledge automatically, but also supports downstream problem routing and schema selection, generating query-relevant evidence to strengthen reasoning and SQL synthesis.

A key distinction from SEED-style approaches is the locus of knowledge construction. SEED-like methods primarily generate evidence on the question side by asking an LLM to infer or "guess" missing evidence. In contrast, CA emphasizes database-side knowledge caching: it mines stable structural/value/semantic knowledge from the database offline, and activates only the necessary parts online via routing. This design is motivated by industrial databases where field semantics are relatively stable while queries are dynamic, making database-side caching more transferable and cost-effective.

We focus on the following research questions:
- **RQ1:** When human evidence is missing, can we construct substitutable evidence purely from database-intrinsic structure and data distributions? How does it compare with gold evidence?
- **RQ2:** Does a multi-agent closed loop ("schema mining → semantic routing → evidence generation") outperform a single evidence generator or schema-only enhancement?
- **RQ3:** What are the prominent limitations or failure modes of CA?

**Contributions**
- We propose Companion Agents, a new Text-to-SQL paradigm that automatically constructs explainable evidence from database-side cached knowledge under missing-evidence conditions.
- We design a collaborative agent system consisting of Schema Mining Agent, Query Routing Agent, and Evidence Generation Agent, enabling offline database profiling and online on-demand evidence construction.
- We demonstrate consistent gains on BIRD under fully missing evidence, significantly narrowing the performance gap between "with evidence" and "without evidence", especially on complex reasoning queries.

## 2. Related Work

**2.1 Text-to-SQL models**

Early Text-to-SQL systems were predominantly rule-based or statistical, relying on syntactic parsing and template matching [12,13], with interactive interfaces such as NaLIR [14]. With deep learning, research shifted toward schema linking, intermediate representations, and pretraining, including sequence-to-sequence frameworks [15] and grammar-/graph-based models [16,17]. However, generalization remained limited by model capacity and training data. Recently, large language models (LLMs) such as GPT-4 [18], Gemini [19], and open-source models such as LLaMA [20] and DeepSeek [21] have driven a new wave of Text-to-SQL research. Prompting-based approaches include C3 [22], DIN-SQL [23], MCS-SQL [24] with self-consistency [25], data synthesis and model scaling efforts such as SENSE [26] and CodeS [9], and self-refinement strategies such as E-SQL [27] and Self-Polish [28]. Fine-tuning and ensemble selection frameworks (e.g., XiYan-SQL [29], CHASE-SQL [30], MSc-SQL [31]) and verification-based systems such as CHESS [7] further improve robustness.

**2.2 Datasets and the role of evidence**

NL2SQL datasets evolved from single-domain corpora (e.g., ATIS [32], GeoQuery [33]) to larger and more complex settings, including domain-specific benchmarks [34–36]. Cross-domain datasets such as WikiSQL [37] and Spider [3] emphasize generalization. BIRD [4] introduces noisy, large-scale databases closer to industrial environments, and categorizes evidence into four types: (i) numeric reasoning knowledge, (ii) domain knowledge, (iii) synonym knowledge, and (iv) value illustration. Importantly, much of this "external knowledge" can be derived from careful analysis of schema, metadata, and data samples—an observation that motivates CA's database-side mining.

## 3. Method

### 3.1 Problem definition

Let a database instance be $D$ with schema $S$ (tables, columns, and foreign-key constraints), an NL question be $q$, and human-provided evidence be $e$. The goal of Text-to-SQL is to generate an SQL query y whose execution result matches the ground truth.

For an LLM-based approach without parameter fine-tuning, SQL generation can be abstracted as:

$$\hat{y} = f_\theta(P(q, S, e))$$

where $f_\theta$ is the LLM, $P(\cdot)$ is a prompt construction function, and e provides additional knowledge related to the question and database. In the evidence-missing scenario ($e=\emptyset$), we propose to construct substitute evidence $\bar{e}$ using Companion Agents:

$$\bar{e} = g_\emptyset(D, S, q), \quad \hat{y} = f_\theta(P(q, S, \bar{e}))$$

Our objective is to maximize execution accuracy (EX) and logical consistency under evidence-missing conditions.

### 3.2 Companion Agents (CA)

The CA framework targets scenarios where schema annotations are missing or incomplete. It proactively mines and generates query-relevant contextual evidence to support robust LLM-based SQL synthesis. CA consists of three core modules (Figure 2):

- **Schema Mining Agent (SMA)**: deep schema mining and database profiling;
- **Query Routing Agent (QRA)**: query-type recognition and evidence-type routing;
- **Evidence Generation Agent (EGA)**: evidence construction, retrieval, and semantic strengthening.

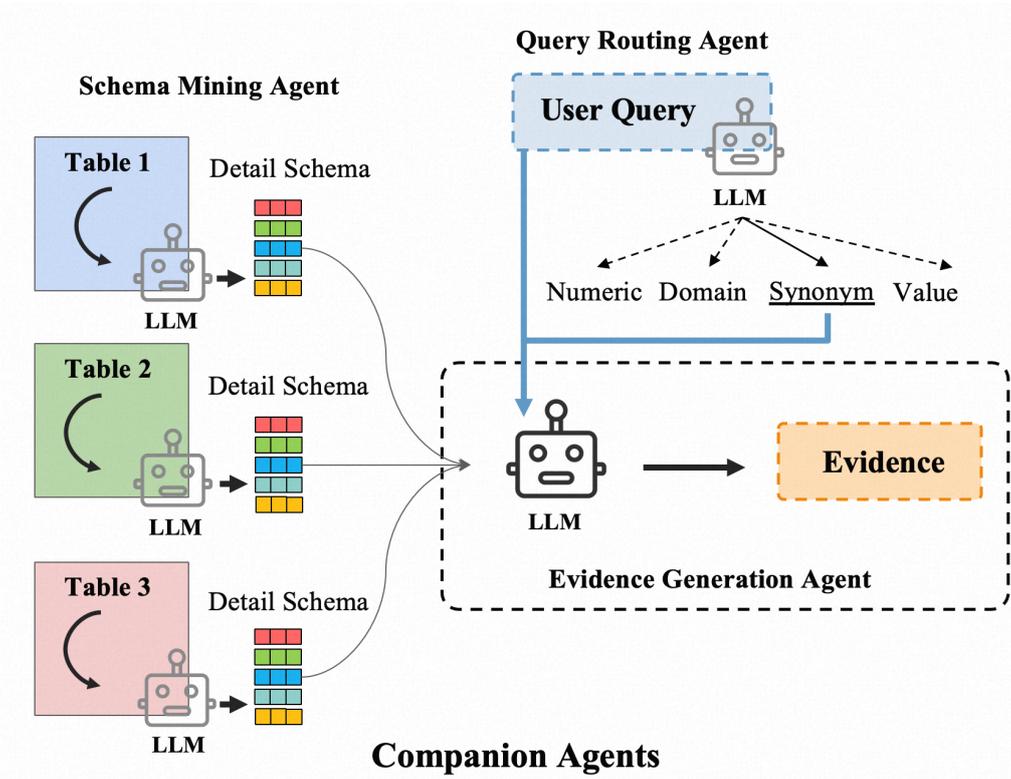

Figure 2. Overview of the Companion Agents framework.

### 3.2.1 Schema Mining Agent: deep structural mining

Based on BIRD's evidence taxonomy, we observe that beyond numeric reasoning formulas, the remaining evidence types—domain knowledge, synonym knowledge, and value illustration—can often be derived from structured schema metadata and sampled data values. SMA is designed to automatically extract such signals and store them as reusable schema-knowledge. It contains three stages:

**(1) Schema Extraction.**

SMA performs automatic structural parsing, statistical profiling, and constrained semantic induction for each database. It first extracts table- and column-level information via database interfaces, including simplified DDL statements, foreign-key constraints, and sampled rows, forming a base structural description. It then computes column-level profiles from samples, including typical values, distribution ranges, and high-frequency enumerations to build reproducible field portraits. Finally, SMA invokes an LLM over (column name + structural context + profile statistics + sampled values) to produce semantic summaries and tags, generating: (i) field semantic descriptions, (ii) candidate aliases/synonym mappings, (iii) time granularity/unit hints, and (iv) readable glossaries for frequent enumerated values. All outputs are encoded into a structured schema-knowledge file for downstream routing and evidence construction.

**(2) Few-shot Knowledge Library Building.**

To enhance reasoning under knowledge scarcity, SMA builds a few-shot QA library and uses an LLM to standardize examples and abstract their structure. Specifically, it extracts (question, SQL) pairs from training data, then asks the LLM to denoise and normalize questions (e.g., resolve references, remove ambiguity) and to skeletonize SQL (e.g., abstract SELECT–FROM–WHERE–GROUP–ORDER templates and operator chains). This produces transferable entries of "semantic question template + SQL logical skeleton". After rule-based deduplication and schema-compatibility checks, SMA yields a clean, reusable QA library that provides experience priors when explicit evidence is missing.

**(3) Similar-case matching.**

Given a new Text-to-SQL query, SMA retrieves the top-k most semantically similar examples from the QA library and concatenates them into a structured few-shot prompt. This enables the LLM to borrow solutions from analogous problems, improving stability on complex reasoning and cross-table join queries.

### 3.2.2 Query Routing Agent: semantic guidance and query typing

Given question q and schema-knowledge, QRA performs multi-label routing (via an LLM) to determine which evidence types are required, including: numeric reasoning, domain knowledge, synonym/alias, and enum/value illustration.

- **Numeric reasoning:** QRA selects candidate numeric columns and uses column profiles (mean/variance/range/quantiles) to build computation-oriented explanation templates that clarify the required fields and reasoning chain.
- **Domain knowledge:** QRA identifies relevant fields using schema-knowledge and schema-linking signals; if database-side signals are insufficient, it can trigger controlled web retrieval as a fallback to supply necessary domain rules.
- **Synonym/alias:** QRA uses SMA-produced alias sets and mapping tables to rewrite/expand the query (synonym substitution, coreference resolution, phrase alignment) to improve column alignment.
- **Enumerations:** QRA extracts value domains and typical samples for discrete fields (e.g., gender, region, category), reducing constraint mismatches and empty-result risks; if enum evidence is needed, QRA provides enum dictionaries to EGA.

### 3.2.3 Evidence Generation Agent: evidence construction and semantic enhancement

EGA generates or retrieves contextual evidence based on QRA's routing. It includes three evidence strategies:

**(1) Schema-consistency evidence.**

By comparing foreign-key constraints and semantic similarity, EGA proposes plausible join relations and connection paths.

Example evidence: "Table customer is linked to order via customer_id, enabling aggregation of order records per customer."

**(2) Semantic constraint evidence.**

EGA generates natural-language constraints for time/range/category conditions.

Example: "Column year denotes the sales year; records with year > 2020 correspond to 2021 and later."

**(3) Logical completion evidence.**

EGA retrieves similar cases from the few-shot library to complete multi-step logic.

Example: "Compute total sales per employee via a subquery, then apply a window function RANK() for group-wise ranking and top-10% filtering."

Finally, EGA embeds these evidence items into the model prompt, forming a closed-loop enhancement pipeline from structure mining and semantic routing to evidence construction.

## 4. Experiments

### 4.1 Datasets

We evaluate on BIRD, which bridges academic and industrial scenarios by introducing large-scale noisy databases. BIRD contains 12,751 NL–SQL pairs spanning 95 databases (33.4 GB) and 37 domains. Queries are categorized into Simple (~60%), Moderate (~30%), and Challenging (~10%), reflecting increasing difficulty.

### 4.2 Evaluation metrics

We use the official BIRD execution evaluation script to compute **Execution Accuracy (EX)**: a prediction is correct if the execution result of the generated SQL exactly matches that of the gold SQL. We report overall EX and EX stratified by difficulty (Simple/Moderate/Challenging).

### 4.3 Baselines

We select representative SOTA baselines with public implementations:
- **RSL-SQL** [5]: robust schema linking + structured prompting.
- **CHESS** [7]: multi-agent candidate generation + unit-test verification.
- **DAIL-SQL** [38]: evidence-heavy prompt engineering / in-context learning.

To isolate method effects, both baselines and CA-augmented versions use the same LLM (**qwen-plus**) for inference.

### 4.4 Implementation details

All methods share the same inference configuration: temperature = 0.2, top-p = 0.9, max_tokens = 1024. For schema extraction, we sample n=200 rows per column (prioritizing distinct values) to build column profiles. Few-shot retrieval uses top-k=5. QRA adopts zero-shot multi-label routing with confidence threshold $\tau=0.5$ to filter irrelevant evidence types. Reported EX values are averaged.

### 4.5 Results

Table 1 summarizes performance under the fully missing evidence setting. Removing human evidence causes substantial degradation for all baselines, especially on Challenging queries, confirming heavy reliance on external annotations. With CA, overall EX improves by 4.49%, 4.37%, and 14.13% on RSL-SQL/CHESS/DAIL-SQL, respectively; on the Challenging subset, gains further increase to 9.65%, 7.58%, and 16.71%. This indicates that CA can effectively mine internal database knowledge and construct substitute evidence, narrowing the gap between "with evidence" and "without evidence".

DAIL-SQL benefits the most because its ICL-style prompting depends most heavily on evidence, whereas RSL-SQL and CHESS already have internal priors (schema linking and unit tests), resulting in milder degradation and smaller but consistent recovery.

Table 1. Execution accuracy without BIRD evidence and recovery with Companion Agents.

|  | Simple | Moderate | Challenging | Total |
| --- | --- | --- | --- | --- |
| RSL-SQL-w/o evi | 61.41 | 40.09 | 29.66 | 51.96 |
| RSL-SQL-evi | 71.68 | 55.60 | 48.97 | 64.67 |
| RSL-SQL-CA | 65.19(↑3.78) | 44.40(↑4.31) | 39.31(↑9.65) | 56.45(↑4.49) |
| Chess-w/o evi | 56.00 | 37.93 | 26.90 | 47.78 |
| Chess-evi | 67.38 | 56.82 | 47.06 | 62.14 |
| Chess-CA | 59.46(↑3.46) | 43.10(↑5.17) | 34.48(↑7.58) | 52.15(↑4.37) |
| DAIL-SQL-w/o evi | 38.76 | 27.39 | 17.77 | 39.52 |
| DAIL-SQL-evi | 63.78 | 51.72 | 40.00 | 57.89 |
| DAIL-SQL-CA | 62.16(↑23.4) | 42.67(↑15.28) | 34.48(↑16.71) | 53.65(↑14.13) |

### 4.6 Ablation study

To quantify each module's contribution, we perform ablations (under the DAIL-SQL baseline) with four variants:
- A) Baseline (w/o evidence)
- B) SMA Only: schema-knowledge only (no routing)
- C) QRA Only: routing only (without substantive knowledge content)
- D) Full CA (SMA + QRA + EGA)

As shown in Table 2, SMA Only provides substantial gains on Simple and Moderate, suggesting that structural/value-domain profiling already compensates for many missing semantics. QRA Only yields limited improvements, indicating that "routing without content" is insufficient. Full CA performs best across all difficulty levels, with the largest gain on Challenging (+16.71), showing that the closed-loop design—knowledge consolidation + on-demand activation + evidence construction—is key to CA's effectiveness.

Table 2. Ablation results on DAIL-SQL.

|  | Simple | Moderate | Challenging | Total |
|---|---|---|---|---|
| w/o evidence | 38.76 | 27.39 | 17.77 | 39.52 |
| SMA only | 55.83 | 36.47 | 27.96 | 48.95 |
| QRA only | 42.14 | 29.82 | 19.61 | 41.39 |
| SMA + QRA(CA) | 62.16 | 42.67 | 34.48 | 53.65 |

## 5. Discussion

Our experiments demonstrate that **Companion Agents (CA)** consistently and significantly improve execution accuracy for three distinct paradigms of Text-to-SQL baselines under evidence-missing conditions, with larger recovery on Challenging queries. This supports the central claim that replacing the conventional "human evidence–dependent" enhancement pathway with a closed-loop mechanism—**database-side knowledge consolidation → on-demand routing → evidence generation**—is both effective and broadly applicable.

To better understand CA's boundary of effectiveness, we manually analyze dev-set failures where CA-augmented systems still produce incorrect SQL. As summarized in Table 3 (from your manuscript), we observe two stable error modes:

1. **Missing enum instantiation.** CA sometimes identifies that a query requires filtering on a categorical/code field, but fails to provide executable value mappings that appear in gold evidence (e.g., mapping "Unified School District" / "Elementary School District" to DOC=54 / DOC=52). Without concrete mappings, EGA cannot instantiate WHERE predicates, leading to missing or incorrect constraints and potentially empty results.

2. **Focus drift from structural rules.** CA may over-attend to secondary semantics and omit structural requirements encoded in gold evidence, such as composing a "full communication address" by selecting Street, City, State, and Zip. Such omissions often manifest as incomplete SELECT fields or mismatched aggregation scope.

These patterns suggest that CA's remaining bottlenecks are not merely whether relevant columns can be found, but whether evidence can be **(i) concretized into executable enum/value mappings** and **(ii) aligned with structural completeness constraints**. In industrial settings with highly discrete enumerations and rich profiles, CA may be further strengthened by integrating an external enum knowledge base for long-tail mappings, and by adding structure-aware coverage checks to ensure that evidence explicitly enumerates required field compositions.

Table 3. Typical failure modes under CA.

| Error Mode | Case ID | NL Query | Gold Evidence | Error Analysis |
|---|---|---|---|---|
| **Incomplete enumeration mapping** | #49 | Compute ratio of **Unified School District** vs **Elementary School District** schools in Orange County | Elementary SD = DOC 52; Unified SD = DOC 54 | States "count and compute ratio" without giving **executable DOC=52/54 mappings**, so WHERE conditions cannot be instantiated |

| Focus drift from structural requirements | #27 | List schools with **full communication address** in Monterey | **Full communication address** includes Street, City, State, and Zip | Diverts to FRPM/age semantics and **omits address-field composition**, leading to incomplete SELECT fields |
|---|---|---|---|---|

# 6. Conclusion

We address the practical challenge of missing evidence and incomplete annotations in industrial Text-to-SQL. We propose Companion Agents (CA), a database-centric paradigm that mines intrinsic structural and value-side signals to build reusable schema-knowledge offline, and constructs query-adaptive evidence online via routing and generation. Under the fully missing evidence setting on BIRD, CA consistently improves multiple SOTA baselines, recovering +4.49 / +4.37 / +14.13 EX points on RSL-SQL / CHESS / DAIL-SQL, with even larger gains on Challenging queries (+9.65 / +7.58 / +16.71). Ablation results further confirm that schema-knowledge caching provides the foundation, while routing and evidence construction determine the upper bound on complex reasoning queries. These findings demonstrate that database-side knowledge caching is a promising and deployable direction for narrowing the gap between academic Text-to-SQL benchmarks and real-world usage.